\documentclass[10pt,journal]{IEEEtran}
\usepackage{times}
\usepackage{helvet}
\usepackage{courier}
\usepackage{latexsym}
\usepackage{amsfonts,amssymb}
\usepackage{amsmath}
\usepackage{graphicx}
\usepackage{algorithm}
\usepackage{algorithmic}
\usepackage{float}
\usepackage{array}
\usepackage{multirow}
\usepackage{url}  
\usepackage{geometry}
\usepackage{setspace}

\title{Knowledge Graph Embedding based on MVF-based Semantic Principle}

\author{Han~Xiao$^*$,
	Yidong~Chen, Xiaodong~Shi
	\IEEEcompsocitemizethanks{\IEEEcompsocthanksitem H.Xiao$^*$, Y.Chen and X.Shi are with the Department of Cognitive Science and Technology, School of Information Science and Engineering, Xiamen University, Xiamen, Fujian 361005, China. \protect 
	\\ $^*$ Corresponding Author: Han Xiao \protect
	\\ $^*$ E-mail: bookman@xmu.edu.cn}
}

\begin{document}

\IEEEtitleabstractindextext{
\begin{abstract}
Knowledge representation is one of the critical problems in knowledge engineering and artificial intelligence, while knowledge embedding as a knowledge representation methodology indicates entities and relations in knowledge graph as low-dimensional, continuous vectors. In this way, knowledge graph is compatible with numerical machine learning models. Major knowledge embedding methods employ geometric translation to design score function, which is weak-semantic for natural language processing. To overcome this disadvantage, in this paper, we propose our model based on multi-view clustering framework, which could generate semantic representations of knowledge elements (i.e. \textit{entities/relations}). With our semantic model, we also present an empowered solution to entity retrieval with entity description. Extensive experiments show that our model achieves substantial improvements against baselines on the task of knowledge graph completion, triple classification, entity classification and entity retrieval.
\end{abstract}

}

\maketitle
\IEEEdisplaynontitleabstractindextext

\section{Introduction}
Knowledge representation is one of the critical and foundational problems in knowledge engineering and artificial intelligence. Traditional knowledge representation methods are logic and symbolic, \cite{sowa1999knowledge}. Thus, they are unsuitable for the trend of deep learning. To facilitate the application of knowledge in statistical learning methods, continuous vectorial representations of entities/relations are necessary. Therefore, knowledge graph embedding (KGE) is studied to fulfill this motivation and benefit many areas such as question answering \cite{Lukovnikov2017Neural} and relation extraction \cite{Wen2018Attention}. 

Specifically, KGE represents a symbolic triple $(h, r, t)$ as real-valued vectors $\mathbf{(h, r, t)}$, each of which corresponds to head entity, relation and tail entity, respectively. Currently, a variety of embedding methods are emerging, including translation-based models such as TransE \cite{bordes2013translating}, neural network based models such as NTN \cite{socher2013reasoning}, probabilistic models such as TransG \cite{xiao2016TransG} etc. 

The embedding framework is abstracted as follow. First, each embedding method proposes a score function of triple $(h, r, t)$, such as $f_r(h,t) = \mathbf{||h+r-t||}_2^2$ for TransE. Then, an objective is constructed by minimizing the score function for the true triples as $\mathcal{L}=\sum_{(h,r,t) \in \Delta} f_r(h, t)$, where $\Delta$ indicates the set of true triples. Last, after the optimization process is converged, the representations are obtained. To summarize, different method branches share the same framework, but differ in the principles of the design of score functions.
 
As a major methodology of knowledge representation, translation-based models (e.g. TransE), metric the error of geometric translation (formally as $\mathbf{h + r \approx t}$) to design score function. There exist many following variants in this branch. Basically, different variants employ different embedding spaces such as hyperplanes \cite{wang2014knowledge}, rotating matrices \cite{lin2015learning} convolution \cite{dettmers2018convolutional} to project the entities. The advantage of this branch is simple and efficient, while the disadvantage is that the models are insufficient to characterize complex knowledge phenomenon such 1-N, N-1, N-N relationship \cite{wang2014knowledge}.

In contrast, neural network based models (e.g. NTN) apply a range of neural architectures as score functions to data-fit the knowledge graph in a black-box style. The simplest method is single layer model \cite{bordes2011learning}, while the current most complex one takes usage of neural tensor network \cite{socher2013reasoning}. The advantage of this branch is to fit knowledge graph in a better degree, while the disadvantage is that this branch only explores some kind of surface-level semantics, i.e, the structure of knowledge graph, rather than the readable meaning in deep level \cite{xiao2017SSP, guo2015semantically}.

Regarding the weak-semantic issue of traditional methods, we take the entity ``Table'' as an example. In TransE, the representation of ``Table'' is a numerical vector as $(0.11, -0.56, 0.98, 0.77, ...)$, from which we can not tell anything semantic such as being furniture, not an animal, etc., because this vector is just a point in high-dimensional geometric space. Similar cases to NTN and other traditional methods. 

In order to jointly achieve the modeling ability and semantic interpretability, probabilistic methods play an important role in knowledge graph embedding. Generative models (e.g. TransG) that take generation distribution as score function could perform effectively with an acceptable semantic interpretability, making a novel way towards knowledge representation \cite{xiao2016TransG}. The advantage of this branch is effective and interpretable, while the disadvantage is that the models are still simple under developing.

Though the probabilistic models have achieved a great success in both the performance and semantic interpretability, this branch still focuses on some specific issues such as multiple relation semantics \cite{xiao2016TransG}, rather than a novel knowledge embedding paradigm. Thus, in this paper, we propose a knowledge graph embedding paradigm based on multi-view clustering framework. We believe a fully developed paradigm could obtain better performance with stronger semantic interpretability. Also, with the semantic representations, natural language processing tasks such as entity retrieval could be empowered extensively. 

\begin{figure}
	\centering
	\label{fig:fig_2}
	\includegraphics[width=\linewidth]{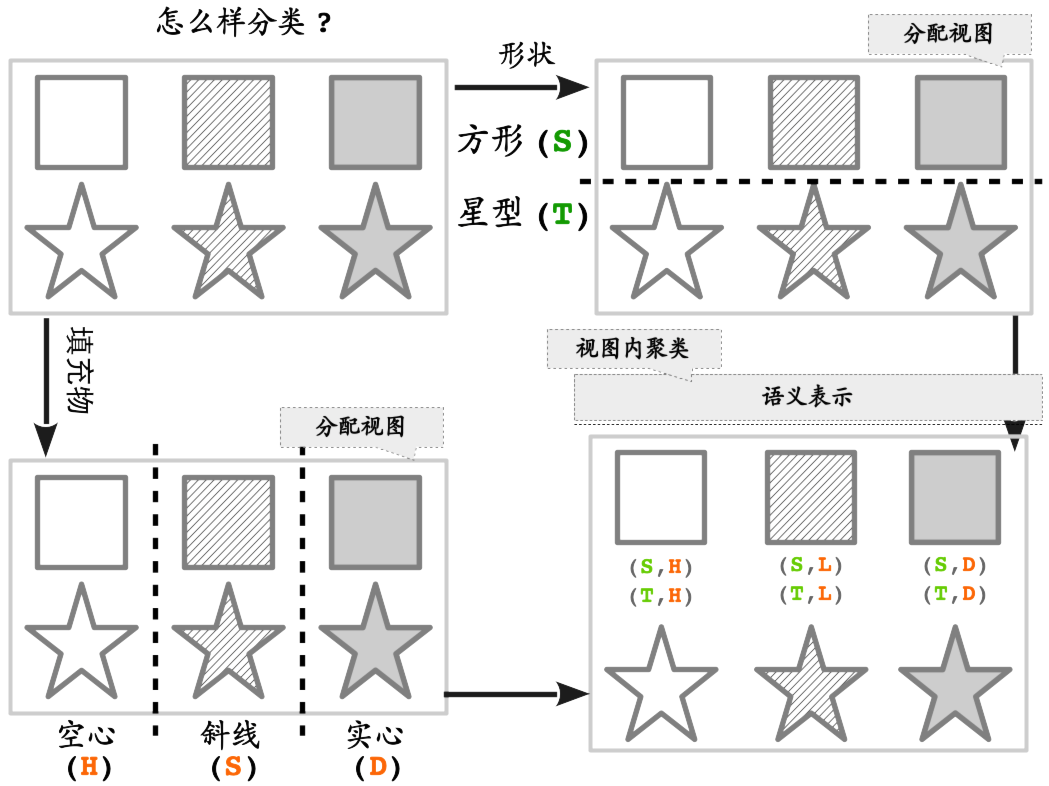}
	\caption{This figure illustrates the principal idea, described in Section 1. Simply, we leverage the cluster ambiguity in the manner of multi-view clustering framework to construct semantic representations.}
\end{figure}

The idea of our paradigm is founded on multi-view clustering formulation, which is motivated by the example shown in Fig.\ref{fig:fig_2}. Referring to Fig.1, there exists a question ``How to categorize these objects?''. Addressed by the \textit{Shape} of objects, two clusters are grouped as ``Square(\textbf{S})'' and ``Star(\textbf{T})'', while focused on the view of \textit{Content}, three clusters are generated as ``Hallow(\textbf{H})'', ``Slash(\textbf{L})'' and ``Solid(\textbf{D})''. In summary, \textit{Shape} and \textit{Content} are the distinguished views for clustering and by gathering the information of clusters in each view, the semantic representations are formed. For the instance in the right-bottom sub-figure, \textbf{(S,D)} indicates a solid \textbf{(D)} square \textbf{(S)}, where the first/second dimension corresponds to the view of \textit{Shape/Content} in a latent manner.

\begin{table*}
	\caption{Different embedding models: score functions $f_r(h,t)$ and model complexity (the number of parameters). $n_e$ and $n_r$ are the number of unique entities and relations, respectively. It is the common case that $n_r \ll n_e$. $k$ is the dimension of embedding space. $s$ is the number of hidden nodes of a neural network or the number of slices of a tensor. $n_f$ is the number of factors.}
	\centering
	\renewcommand\arraystretch{1.3}
	\begin{tabular}{c|c|c}
		\hline 
		Model & Score Function $f_r(h, t)$ & \# Parameters \\ 
		\hline 
		\hline TransE \cite{bordes2013translating} & $||\mathbf{h + r - t}||_{l_{1/2}}, \mathbf{r}\in\mathbb{R}^k$ & $\mathcal{O}(n_ek + n_rk)$ \\ 
		\hline Distant Model \cite{bordes2011learning} & $||W_{rh}\mathbf{h} - W_{rt}\mathbf{t}||_1, W_{rh}, W_{rt} \in \mathbb{R}^{k \times k}$ & $\mathcal{O}(n_ek + 2n_rk^2)$ \\
		\hline Bilinear Model \cite{jenatton2012latent} & $\mathbf{h}^T W_r \mathbf{t}, W_r \in \mathbb{R}^{k \times k}$ & $\mathcal{O}(n_ek + n_rk^2)$ \\
		\hline Single Layer \cite{socher2013reasoning} &
		\begin{tabular}{c}
			$\mathbf{u}_r^Tf(W_{rh}\mathbf{h} + W_{rt}\mathbf{t} + \mathbf{b}_r)$ \\
			$\mathbf{u}_r, \mathbf{b}_r \in \mathbb{R}^s, W_{rh}, W_{rt} \in \mathbb{R}_{s \times k}$
		\end{tabular} & $\mathcal{O}(n_ek + n_r(sk + s))$ \\
		\hline NTN \cite{socher2013reasoning} & 
		\begin{tabular}{c}
			$\mathbf{u}_r^Tf(\mathbf{h}^T W_r \mathbf{t} + W_{rh}\mathbf{h} + W_{rt}\mathbf{t} + \mathbf{b}_r)$ \\
			$\mathbf{u}_r, \mathbf{b}_r \in \mathbb{R}^s, W_{rh}, W_{rt} \in \mathbb{R}_{s \times k}, W_r \in \mathbb{R}^{k \times k \times s}$
		\end{tabular} & $\mathcal{O}(n_ek+n_r(sk^2+2sk+2s)$ \\
		\hline TransH \cite{wang2014knowledge} & 
		\begin{tabular}{c}
			$||\mathbf{(h - w_r^Thw_r) + d_r - (t - w_r^Ttw_r)}||_2^2$ \\
			$\mathbf{w_r, d_r} \in \mathbb{R}^k$
		\end{tabular} & $\mathcal{O}(n_ek + 2n_rk)$ \\
		\hline TransR \cite{lin2015learning} & $||\mathbf{M_rh + r - M_rt}||_{l_{1/2}}, \mathbf{M_r} \in \mathbb{R}^{k \times k}$ & $\mathcal{O}(n_ek + n_rk + n_rk^2)$ \\
		\hline TransA \cite{xiao2015TransA} & $\mathbf{(h+r-t)^T W_r (h+r-t)}, \mathbf{W_r} \in \mathbb{R}^{k \times k}$ & $\mathcal{O}(n_ek + n_rk + n_rk^2)$ \\
		\hline TransG \cite{xiao2016TransG} & $\sum_{i=1}^{n_f} \alpha_i e^{-||\mathbf{h + r -t}||}$ & $\mathcal{O}(n_f(n_ek + n_rk + n_r))$ \\
		\hline KG2E \cite{he2015learning} & $\mathbf{\mu^T\Sigma^{-1}\mu} + log det \mathbf{\Sigma}, \mathbf{\Sigma} = \mathbf{\Sigma_h + \Sigma_t + \Sigma_r} $ & $\mathcal{O}(2k_en_e+2k_rn_r)$ \\
		\hline HolE \cite{nickel2015HOLE} & $\sigma(\mathbf{r}^T(\mathbf{h} \otimes \mathbf{t}))$, $\otimes$: convolutional operator& $\mathcal{O}(n_ek+n_rk)$ \\
		\hline ConvE \cite{dettmers2018convolutional} & $f(vec(f(\mathbf{[\overline{h}, \overline{r}]} * \omega)) \mathbf{W})\mathbf{t}$ & $\mathcal{O}(n_ek + n_ek')$ \\
		\hline ProjE \cite{shi2017proje} & $g(\mathbf{W^c}f(\mathbf{e} \oplus \mathbf{r}) + b_p) $ & $\mathcal{O}(n_ek + n_rk + 5k)$ \\
		\hline 
	\end{tabular} 
\end{table*}

Inspired by the above principal paradigm, we leverage a two-level hierarchical generative process for semantic knowledge representation. Fig.2 illustratively exemplifies the generative process of our model. First, the first-level process generates many knowledge views with different semantics, such as \textit{University} and \textit{Location}. The number of the views is decided by the hyper-parameter, while the semantics of the view is decided by the feedback of the second-level process. Then, the second-level process groups the entities/relations/triples, according to the corresponding semantic views. Last, summarizing the cluster identification within each view, KSR constructs the semantic representation of knowledge elements. For the example of \textit{Tsinghua University}, the \textit{Yes} cluster is assigned to \textit{University} view, while the \textit{Beijing} cluster is assigned to \textit{Location} view. By exploiting the multi-view clustering form, knowledge is semantically organized, as \textit{Tsinghua University = (University:Yes, Location:Beijing)}. 

\begin{figure}
	\centering
	\label{fig:fig_3}
	\includegraphics[width=\linewidth]{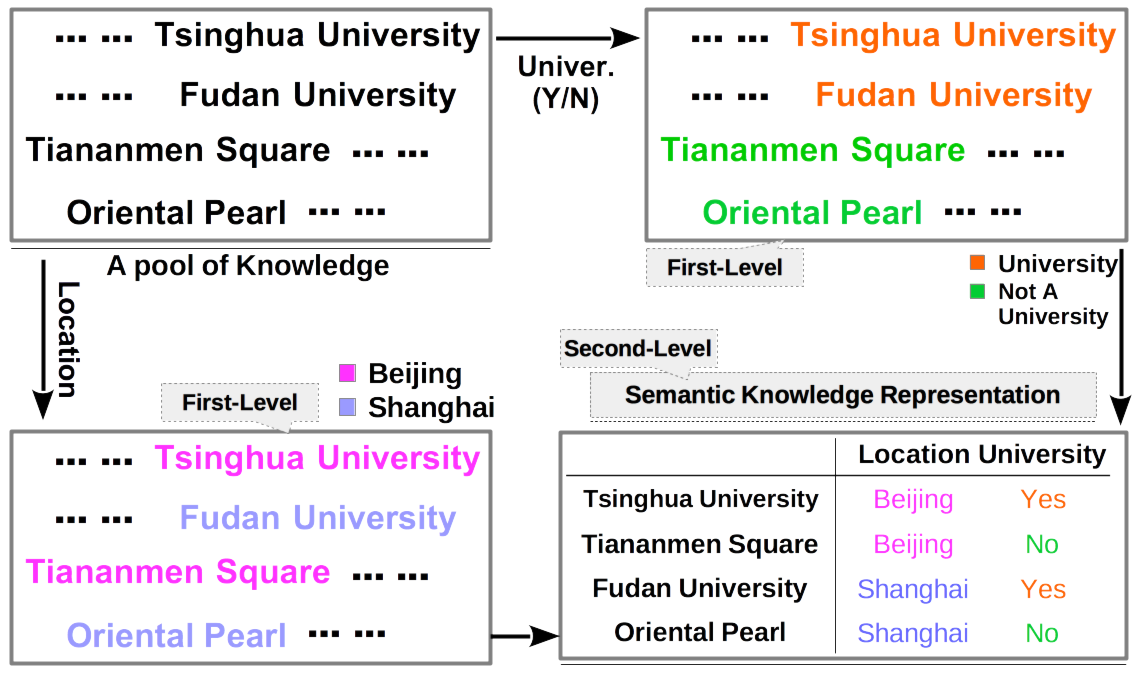}
	\caption{This figure demonstrates the generative process of KSR from the clustering perspective. The original knowledge is semantically clustered from multiple views. Specifically, knowledge views such as \textit{Location}, are generated from the first-level generative process, denoting the ``type of the clusters''. The cluster such as \textit{“Beijing”} in each knowledge view, is generated from the second-level generative process.}
\end{figure}

It is noted that, cluster identification could be in various forms, and in this model, we leverage the membership degree for clusters (i.e. \textit{probabilistic distribution}) to identify entity/relation similar to LDA \cite{blei2012probabilistic},  rather than the linguistic lexicons (e.g. \textit{Yes, Beijing,} etc.). Thus, our semantic representation is a vector of concatenated probabilistic distributions, rather than a list of lexicons. In other words, all the knowledge views (e.g. \textit{University}) are latent concepts as the topics in LDA \cite{blei2012probabilistic}. However, with the aid of textual description of knowledge graph, we can easily map the latent views and clusters into the human-readable words.

Actually, there exists the textual description for each entity in mainstream knowledge graph (e.g. Freebase \cite{bollacker2008freebase}). For example, the description of entity ``Artificial Intelligence'' is ``Artificial intelligence is the intelligence exhibited by machines or software''. For the task of entity retrieval, we jointly embed the knowledge elements (i.e \textit{entities/relations}) and textual descriptions into the same semantic space. Then, given a query as a word sequence, we could predict the corresponding entity by semantic matching.

\textbf{Experiments.} We conduct our experiments on the subsets of Freebase for the task of entity classification, knowledge graph completion, semantic analysis and entity retrieval. Experimental results on these datasets demonstrate that our model outperforms the other baselines with remarkable improvements.

\textbf{Contributions.} Our contributions are three folds:
\begin{enumerate}
	\item Based on the multi-view clustering methodology, we have provided a novel unsupervised paradigm for semantic representation of knowledge graph.
	\item Based on our semantic representations, we have proposed several methods for entity retrieval, which offers a potential application for jointing knowledge and language. 
	\item Experimental results illustrate the effectiveness of our theoretical analysis and proposed methods.
\end{enumerate}

\textbf{Organization.} In Section 2, we survey the related researches and categorize them into lines. In Section 3, we introduce our knowledge graph embedding methodology within multi-view clustering framework. In Section 4, we design a joint model of knowledge and language for the task of entity retrieval. In Section 5, we conduct experiments for our models and carry out a case analysis for our semantic representations. 

\section{Related Work}
We list the score functions of different methods in Tab.1, the details of which are followed. 

To numerically represent knowledge elements (i.e. \textit{entities/relations}), distant model \cite{bordes2011learning} is proposed. It introduces two independent projections to the entities for a specific relation. This model starts the neural network based research trend but there are two disadvantages for this method. First, distant model is weak in capturing correlations between entities and relations as it uses two separate matrices. Thus, bilinear model \cite{jenatton2012latent} applies the quadratic form between entity embeddings to characterize the correlations. Second, distant model is weak in data-fitting ability. Thus, single layer model \cite{socher2013reasoning} introduces nonlinear transformations by neural network to fit the knowledge triples. To overcome both the disadvantages, NTN \cite{socher2013reasoning} jointly applies the second-order form and neural layers, making a state-of-the-art method. The advantage of this branch is to model knowledge in a better degree, while the disadvantage is lack of semantic interpretability. 

To further tackle this task in a simple and effective manner, TransE \cite{bordes2013translating} is proposed as a pioneering work of the translation-based methods, which translate the head entity to the tail one by the relation vector, formally as $\mathbf{h + r \approx t}$. This method starts to embed knowledge within geometric principle. However, there exist two issues for TransE. First, there are many tail/head entities for one head/tail entity under the specific relation in knowledge graph, which is denoted as complex relation issue. In the other word, from the perspective of translation, one head entity would be translated to many distinguished tail entities by the same relation vector or many head entities would be translated to one tail entity by the same relation vector, no matter of which could trigger an imprecise issue. To tackle this problem, TransH \cite{wang2014knowledge} is proposed to project the relation-specific head and tail entities into the same hyperplane, then perform the translation-based embedding process. In this way, the complex relation issue could be alleviated. Second, the entity space in TransE is identical without any transformation, which leads to a very weak data-fitting model. To strengthen the data-fitting ability, many variants are proposed. TransR \cite{lin2015learning} rotates the entity space with a relation-specific matrix, while CTransR \cite{lin2015learning} performs such operations on the clustered entities rather the entire entity space. TransA \cite{xiao2015TransA} employs the idea of metric learning to plug a relation-specific metric matrix into the distance measurement. TransD \cite{Ji2016Knowledge} follows the work of TransR and it uses two vectors to identify a knowledge element (i.e. \textit{entity/relation}), where the first one represents the meaning of a(n) entity (relation) and the other one is used to construct mapping matrix dynamically. TransM \cite{fan2014transition} also enhances the relation-specific transformation in TransR with a scalar. Besides, convolutional operator is also applied in this branch, \cite{dettmers2018convolutional}.  The advantage of this branch is simple and efficient, while the disadvantage is that the models are insufficient to characterize complex knowledge phenomenon.  

In order to jointly achieve the modeling ability and semantic interpretability, TransG \cite{xiao2016TransG} as a probabilistic method is proposed. This method generates the representations of entities and relations from a Gaussian mixture model to fix the issue of multiple relation semantics, which indicates that there exist many semantics for a specific relation.
  
Further researches incorporate additional structural information into embedding. PTransE \cite{lin2015modeling} takes the advantages of entity path in a composition manner and achieves the state-of-the-art performance. \cite{wang2015knowledge} incorporates the heuristic rules into embedding and provides a new perspective for leveraging external information. SSE \cite{guo2015semantically} takes full advantage of additional semantic information and enforce the embedding space to be semantically smooth, i.e., entities belonging to the same semantic cluster will lie close to each other in the embedding space. Also, KG2E \cite{he2015learning} involves Gaussian analysis to characterize the uncertain concepts of knowledge graph. \cite{zhong2015jointly} aligns the knowledge graph with the textual corpus and then jointly conduct the knowledge and word embedding. SSP \cite{xiao2017SSP} extends the translation-based embedding methods from the triple-specific manner to the “Text-Aware” fashion by encoding textual descriptions of entities.  There are also some other work such as ManifoldE \cite{xiao2016ManifoldE}, HOLE \cite{nickel2015HOLE}, SE \cite{bordes2011learning, socher2013reasoning} and RESCAL \cite{nickel2012factorizing}, \cite{shi2018open} etc.

\begin{figure*}  
	\begin{small}
		\begin{eqnarray}
		\label{eqn:3}
		& [h,r,t,\{z_k, y_k\}_{1...n}, f, \sigma] & = \prod_{k=1}^n [z_k|h][z_k|r][y_k|t][y_k|r][z_k, y_k, f^{=k}, \sigma] \\
		\label{eqn:1}
		& [h,r,t,\{z_k, y_k\}_{1...n} | f, \sigma] &= \prod_{k=1}^n [z_k|h][z_k|r][y_k|t][y_k|r][z_k, y_k | f^{=k}, \sigma] \\
		& [h,r,t] & = \sum_{k=1}^n \sum_{i_k, j_k=1}^d [h, r, t, \{z_k^{=i_k}, y_k^{=j_k}\}|f^{=k}, \sigma] \\
		& & = \prod_{k=1}^{n} [f^{=k}|\sigma] \sum_{i_k, j_k=1}^d [h,r,t, {z_k^{=i_k}}, {y_k^{=j_k}} | f^{=k}, \sigma] \\
		& & = \overbrace{\prod_{k=1}^{n} [f^{=k}|\sigma] \overbrace{\left\{ \sum_{i_k,j_k=1}^{d} [z_k^{=i_k}|h][z_k^{=i_k}|r][y_k^{=j_k}|t][y_k^{=j_k}|r][z_k^{=i_k}, y_k^{=j_k}| f^{=k}, \sigma] \right\}}^{Second-Level~Mixture}}^{First-Level~Mixture} \label{eqn:2} ~^{\spadesuit} \\
		& & = \prod_{k=1}^{n} [f^{=k}|\sigma] \sum_{i_k,j_k=1}^{d} [z_k^{=i_k}|h][z_k^{=i_k}|r][y_k^{=j_k}|t][y_k^{=j_k}|r]e^{\frac{|[z_k^{=i_k}|h][z_k^{=i_k}|r] - [y_k^{=j_k}|t][y_k^{=j_k}|r]|}{\sigma}} ~^{\spadesuit} \nonumber \\ \label{eqn:5} 
		\end{eqnarray} 
	\end{small}
\end{figure*}

\section{Methodology}
Overall, the model in embedding framework should indicate the score function $f_r(h,t)$, which is achieved as a probabilistic distribution $\mathcal{P}(h, r, t)$ in our method by the generative process described in Section 3.1. The mathematical details stem from the generative process in a conventional Bayesian modeling manner. Notably, as some statistical literature introduced, for brevity, we replace $ \mathcal{P}(a) \doteq [a], \mathcal{P}(a | b) \doteq [a | b]$.

\subsection{Model Description}
We leverage a two-level hierarchical generative process to semantically represent the knowledge elements (i.e. \textit{entities/relations}) as follows:

\begin{figure}[H]
		\textit{
			\parbox{0.95\linewidth}{
				For each triple (head, relation, tail) as $(h, r, t) \in \Delta$
				\\
				\textbf{(First-Level)} \\ 
				For each knowledge view $f_i$:
				\begin{enumerate}
					\item \textbf{(Second-Level)} \\  Draw a subject-specific cluster $z_i$ from $$ [z_i] \propto [z_i | h] [z_i | r] [z_i | t, f_i] $$
					\item \textbf{(Second-Level)} \\ Draw an object-specific cluster $y_i$ from $$ [y_i] \propto [y_i | t] [y_i | r] [y_i | z_i, f_i]$$
				\end{enumerate}
	}}
\end{figure}

Specifically, the first-level generation process produces the clustering views. In detail, we just allocate the different views in the first-level process, according to the hyper-parameter $n$, which means if there are $10$ views (the hyper-parameter is as $n=10$), we should allocate $n=10$ views in the first level process. Initially, the views are not human comprehensive, because the clusters in each view are not formed. After the clustering process converges, the views are finally semantic according to the distribution of the clusters in the corresponding view.

In the above process, $\Delta$ is the set of golden triples, which means our training dataset. \textit{\textbf{All the parameters of $[f_i]$, $[z_i|h]$, $[z_i|r]$, $[y_i|t]$, $[y_i|r]$ are learned by the training procedure}}, and $[h]$, $[r]$, $[t]$ are uniformly distributed, indicating that they can be safely omitted with simple mathematical manipulation.

With the generative process, we could draw the probabilistic graph as shown in Fig.3. For example, $y_n$ is based on $t, r, f_n, z_n$ in the generative process, then there exist four links from condition variables $t, r, f_n, z_n$ to the generated variable $y_n$. However, the edge in probabilistic graph implies the independent relationship. Thus, we could work out the joint probabilistic distribution (Equation (\ref{eqn:3})), according to the probabilistic graph as shown in Fig.3. Then, with the Bayesian rule, we could work out the conditional distribution as Equation (\ref{eqn:1}). Last, with the sum rule \cite{murphy2012machine}, the score function is worked out, where $n$ and $d$ are the number of views and clusters, respectively. \textbf{\textit{Notably, the generative probability $[h, r, t]$ of the triple $(h, r, t)$ as Equation (\ref{eqn:2}) is our score function.}} The number of clusters in different views is fixed as the hyper-parameter $d$. Notably, the variable $k$ means $k$-th view in the Equation (\ref{eqn:1}-\ref{eqn:5}).

\begin{figure}
	\centering
	\includegraphics[width=\linewidth]{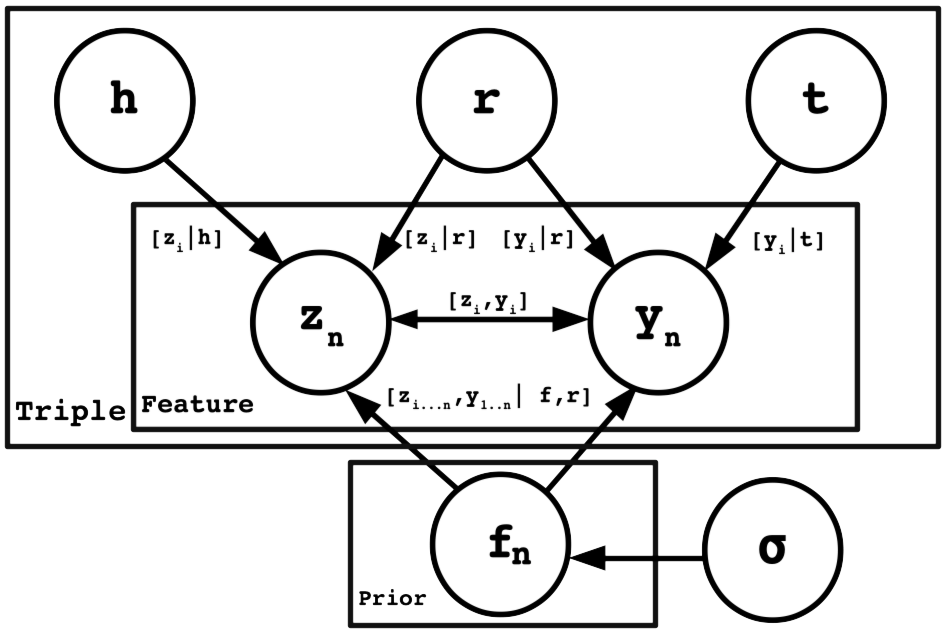}
	\label{fig:fig_1}
	\caption{The probabilistic graph of the generative process. The outer plate corresponds to the first-level and the inner one corresponds to the second-level. The specific form of each factor is introduced in Section 3. $h, r, t$ denote the head entity, relation and tail entity, respectively. $f_n$ denotes the $n$-th view and $\sigma$ is the hyper-parameter of the prior distribution. $z_n$ is subject-related cluster id in the $n$-th view, while $y_n$ is the object-related cluster id in the $n$-th view.}
\end{figure}

Regarding the representation transformation from generation process to the probabilistic graph, we modify a generation detail. First, as we know, $z_n$ and $y_n$ are mutually affected, which is reasonable, because subject and object always affect each other. Thus, $z_n$ and $y_n$ are bi-directional in Fig.3. Corresponding to the generation process, it is better to apply that $[z_i]$ stems from $[z_i|y_i, f_i]$ rather than $[z_i|t, f_i]$. But in this way, there would be a loop for the generation process. Thus, we approximate $[z_i|y_i, f_i]$ with $[z_i|t, f_i]$ because both $y_i$ and $t$ are object-related.

To illustrate the process of probabilistic graph model, we take Fig.2 as an example. Regarding the triple of \textit{(Tsinghua University, Friends, Fudan University)}, in the first-level generation, there are initially two views, according to the hyper-parameter $n=2$. We would process the clustering procedure for the two views in the second-level generation. For the first view that ``Location'', our model draws ``Beijing'' cluster from the distribution $[z_i] \propto [z_i|h][z_i|r][z_i|t, f_i]$ for the head entity ``Tsinghua University'' and ``Shanghai'' cluster from the distribution $[y_i] \propto [y_i|t][y_i|r][y_i|z_i, f_i]$ for the tail entity ``Fudan University'', in the second-level process. For the second view that ``University (Y/N)'', our model draws ``Yes'' cluster from the distribution $[z_i]$ for the head entity and ``Yes'' cluster from the distribution $[y_i]$ for the tail entity in the second-level process. Thus, for ``Tsinghua University'', it belongs to ``Beijing'' cluster in the view of ``Location'' and ``Yes'' cluster in the view of ``University(Y/N)''.

It is natural to adopt the most possible cluster in the specific knowledge view as the semantic representation. Suggested by the probabilistic graph (Fig.3), the exactly inferred representation for an entity $S_e=(S_{e, 1}, S_{e, 2}, ..., S_{e, n})$ or a relation $S_r=(S_{r, 1}, S_{r, 2}, ... ,S_{r, n})$ is 
\begin{eqnarray}
S_{e, i} = \arg \max_{c=1}^{d} [z_i = c|e] \\
S_{r, i} = \arg \max_{c=1}^{d} [z_i = c|r][y_i = c |r] 
\end{eqnarray}

In the sequel, we will discuss the form of each distribution factor, respectively.

\let\thefootnote\relax\footnotetext{$^\spadesuit$ $z_k^{=i}$ is short for $z_k = i$, and it is similar for other cases.}

\subsubsection{Regarding Head-/Tail-Specific Cluster: $z_i, y_i$}
For a single entity $e$, the head- and tail-specific cluster is consistent, mathematically $[z_i|e] = [y_i|e]$. Because, no matter the entity \textit{(e.g. Stanford University)} is a subject or an object, the corresponding semantics is identical. Also, it is noteworthy that the terms involved with relations are distinguished for being subject or object, namely $[z_i|r] \neq [y_i|r]$. For the example of triple $(Shakespeare, Write, Macbeth)$, the head-specific cluster $[z_i|r]$ means the active form of ``Shakespeare writes'' while the tail-specific cluster $[y_i|r]$ means the passive form of ``Macbeth is written'', which is a significant difference.

\subsubsection{Regarding $[z_i|y_i, f_i]$ or $[z_i, y_i | f_i]$}
\textit{\textbf{Since a triple is too short to imply more facts, the consistent assumption between head and tail entity is reasonable. The consistent assumption supposes that the semantics between head and tail entity should be generally proximal}}. For the example of triple \textit{(Yangtze River, Event, Battle of Red Cliffs)}, if the head entity suggests the subject-specific location view samples the cluster of \textit{China} with probability \textit{95\%} ($[z_{location} = China] = 0.95$) and \textit{America} with \textit{5\%} ($[z_{location} = America] = 0.05$), then the tail is supposed to suggest the object-specific view samples \textit{China} cluster with much higher probability than \textit{America}, $[y_{location} = China] \gg [y_{location} = America]$. We expect the head and tail could tell one exact story, so we should guarantee the coherence between the sampling distributions. Thus, a Laplace prior is imposed to approximate the distributions of head and tail, or mathematically: $[z_i|t, f_i, \sigma] \propto \exp(-\frac{|[z_i] - [y_i|t]|}{\sigma})$, $[y_i|z_i, f_i, \sigma] \propto \exp(-\frac{|[z_i] - [y_i]|}{\sigma})$, where $\sigma$ is the hyper-parameter of Laplace Distribution and $[z_i]$, $[y_i]$ are presented in the generative process. \textit{\textbf{Notably, we encourage the semantics between head and tail entity consistent in most views generally, which means there always exist some views, in which, the semantics of head and tail are different.}} This fact is reasonable, because if all the semantics of head and tail are identical, our model cannot distinguish the head and tail entities. 

There could be many distributions suitable for this prior, such as Gaussian distribution, Student distribution, etc. But we suppose the the gap between semantics distribution should be sparse. For example, there could be few dimensions, in which, the gap is obvious, while for other dimensions, the gap should approach to zero $0$. For the reason of sparsity, we apply the Laplace prior.

\subsection{Objective \& Training}
To be compatible with the embedding framework, maximum data likelihood principle rather than traditional sampling methods is employed for training. Thus, we maximize the ratio of likelihood of the true triples to that of the false ones, conventionally. Our objective is as follows:
\begin{eqnarray}
\mathcal{L} = \sum_{(h,r,t)\in \Delta} ln [h, r, t] - \sum_{(h',r',t')\in \Delta'}  ln [h', r', t'] 
\end{eqnarray}
where $\Delta$ is the set of golden triples that is the training dataset and $\Delta'$ is the set of false triples, generating from negative sampling \cite{wang2014knowledge}. The specific formula of $[h, r, t]$ as the likelihood is presented in the previous subsection (Equation (\ref{eqn:2})) and all the distribution parameters (i.e.$[f^{=k}|\sigma], [z_k^{=i}|h], [z_k^{=i}|r], [y_k^{=i}|t], [y_k^{=j}|r]$) are learned by the optimization process of SGD to maximize the target $\mathcal{L}$. This training procedure is very similar to that in \cite{xiao2016ManifoldE}.

As to the efficiency, theoretically, the time complexity of our training algorithm is $O(nd^2)$ where $n$ is the view number and $d$ is the cluster number for each view. If $nd^2 \approx d'$ where $d'$ is the embedding dimension of TransE, our method is comparative to TransE in terms of efficiency, while this condition is practically satisfied.  In the real-word dataset FB15K, regarding the training time, TransE costs \textit{11.3m} and KSR costs \textit{23.4m}, which is almost the same. \textit{The practical time differences stem from our parallel framework (OpenMP Intel Compiler 2017), which optimizes simple models, better.} From our experience of this parallel framework, we suppose the results from FB15K are representative for different data scalability, which means for larger datasets such as DBPedia, the time difference under this parallel framework between KSR and TransE is also twice. Besides, to speed up our method, we can also leverage Hadoop or Spark, which may optimize our method and TransE nearly in the same degree and reduces the time gap between the two methods. Also, for a comparison, in the same setting, TransR needs \textit{485.0m} and KG2E costs \textit{736.7m}. Note that TransE is almost the fastest embedding method, which demonstrates that our method is nearly the most efficient.

\subsection{Analysis from the Identification Perspective (Focus on Performance)}
The plausibility of triples in our model could be discriminated much better. Firstly, in the second-level, the false triple has a low probability for being assigned to any cluster. Secondly, in the first-level, even if some view of this negative one holds high certainty, the corresponding relation also weights the views with $[f_i|\sigma]$ to filter out these noisy information. To summarize, our model could discriminate the plausibility of triples in a two-level filtering form, leading to a better performance. 

\subsection{Analysis from the Clustering Perspective (Focus on Comprehensibility)}
Essentially, regarding the mixture form of Equations (\ref{eqn:1}) and (\ref{eqn:2}), our method takes the spirit of mixture model at both first- and second-level, which could be further analyzed from the clustering perspective. The second-level generative process clusters the knowledge elements (i.e. \textit{entities/relations}) according to a specific knowledge view. These views stem from the first-level process, mathematically according to all the probabilistic terms involved with $f^{=i}$. Furthermore, the first-level generative process adjusts different knowledge views with the feedback from the second-level. Mathematically, the feed-back corresponds to $[z_{1...n}, y_{1...n}, f | h, r, t]$. \textbf{\textit{In essence, knowledge is semantically organized in a multi-view clustering form, Thus, by modeling the multi-view clustering nature, KSR is semantically interpretable.}}

Notably, each view corresponds to a kind of semantics. According to Fig.2, the knowledge semantics are generated from the first-level process, which means discovering different views for clustering. Then, in each knowledge view, a specific cluster is assigned for every entity/relation/triple, which is the second-level process. Last, summarizing the cluster identification within each view, KSR constructs the semantic representation of knowledge elements. For the example of Fig.2, \textit{Tsinghua University} belongs to \textit{Yes} cluster in \textit{University} view and belongs to \textit{Beijing} cluster in \textit{Location} view. In summary, our model represents this entity semantically, as \textit{Tsinghua University = (University:Yes, Location: Beijing)}.

Generally, there are at least two aspects for interpretability: making the model controllable for human and showing the semantics for achieved representations. Our proposed method KSR belongs to the second aspect that to show the semantics for achieved representations.

\section{Entity Retrieval}
Actually, there exists the textual description for each entity in mainstream knowledge graph such as Freebase. For example, the description of entity ``Artificial Intelligence'' is ``Artificial intelligence is the intelligence exhibited by machines or software''.

Motivated by the aligned corpus, we propose a method for the task of entity retrieval in this section. Actually, our knowledge embedding method sets up a semantic space, where entities are differentiated as the concatenation of probabilistic distribution vectors namely $[z|e] = ([z_1|e], [z_2|e], ...,[z_i|e],...[z_n|e])$. Thus, by projecting the words in entity description into the semantic space of corresponding entity, we could work out the semantic representation for each individual word. Then, given a sentence, the semantic representation of the query could be formulated in three different proposed manners. Last, our method metrics the representations of the query and each entity to propose the most possible answer. 


For an entity $e$, there exist the textual description as word sequence $(w_{e,1}, w_{e,2}, w_{e,3}....w_{e,m})$ and semantic embedding distribution $[z|e]$. Here, we have two notes. First, the head- and tail-specific semantic clusters for an entity are identical $[z|e] = [y|e]$ as discussed in Section 3.1.1. Second, a word $w$ could appear in the descriptions of many entities, which we annotate as $\{e_{w,1}, e_{w,2}, .... e_{w, t_w}\}$. Thus, the semantic representation of a specific word $w$ is the average of the semantic representations of co-occurring entities, mathematically as:
\begin{eqnarray}
[z | w] = \frac{\frac{1}{t_w} \sum_{i=1}^{t_w} [z| e_{w, i}]}{\sum_{j=1}^n\frac{1}{t_w} \sum_{i=1}^{t_w} [z^{=j}| e_{w, i}]}
\end{eqnarray}
where $t_w$ is the number of co-occurring entities for the specific word $w$ and the denominator makes the formula a distribution. Actually, the idea behind our method stems from the average pooling, which is a conventional technique in neural natural language processing, \cite{Wang2017Bilateral}. 

For a query $q$ as sequence $q = (w_{q,1}, w_{q,2}, ... w_{q,m})$, we propose three manners to construct the composition: average pooling, naive Bayesian composition and LSTM (long short term memory neural network).

\textbf{Average Pooling} is a conventional technique to compose a set of vectors. The underlying idea is simple and it often makes acceptable performance. Mathematically, the semantic representation is calculated as:
\begin{eqnarray}
	[z | q] = \frac{\frac{1}{m} \sum_{i=1}^m [z|w_{q,i}]}{\sum_{j=1}^d \frac{1}{m} \sum_{i=1}^m [z^{=j}|w_{q,i}]}
\end{eqnarray}
where the numerator is the average pooling and the denominator makes this formula a distribution.

\textbf{Naive Bayesian Composition} takes the assumption of Naive Bayes, which means each individual word makes an independent and equivalent effect on the query, as 
\begin{eqnarray}
	[z | q] = \frac{\prod_{i=1}^m [z|w_{q,i}]}{\sum_{j=1}^d \prod_{i=1}^m [z^{=j}|w_{q,i}]}
\end{eqnarray}
where the numerator is the Naive Bayes product, and the denominator is the normalized factor, which makes $[z_i | q]$ as a distribution.
 
\textbf{LSTM} is a component of deep learning and often deals with sequential data, \cite{Greff2017LSTM, Sundermeyer2012LSTM, Huang2015Bidirectional}. We treat each entity description as an item of training data, where the input word representation is $[z|w]$ and the final hidden representation of LSTM is treated as the output which is labeled as $[z|e]$.  Then, we take mean square error as the objective to train the LSTM. Regarding the inference process:
\begin{eqnarray}
 [z | q] = LSTM([z|w_{q,1}], ..., [z|w_{q, m}])
\end{eqnarray}
 
Given the semantic representation $[z | q]$ of the query $q$, we compare the semantic representation of each entity in knowledge graph with KL diversity:
\begin{eqnarray}
S_{e,q} = KL([z | q] || [z | e])
\end{eqnarray}
where $S_{e,q}$ is the matching score and $[z|q]$/$[z|e]$ is the semantic representation of query/entity. The answer entities are come up with the best matching scores.
\begin{eqnarray}
e_{answer, q} = \arg \min_e S_{e,q}
\end{eqnarray}

\section{Experiments}
In this section, we first introduce the basic experimental settings. Second, three classical experiments are conducted to verify our model performance. Third, to further demonstrate our model is semantically interpretable, we carry out semantic analysis experiments. Last, two extra experiments demonstrate the effectiveness of our entity retrieval model.

\subsection{Experimental Settings}
\textbf{Datasets.} Our experiments are conducted on public benchmark datasets that are the subsets of Freebase \cite{bollacker2008freebase} and Wordnet \cite{miller1995wordnet}. The entity descriptions of FB15K are the same as DKRL \cite{xie2016DKRL}, each of which is a small part of the corresponding wiki-page. The statistics is listed in Tab.\ref{tab7}.

\begin{table}[H]
	\centering
	\caption{Statistics of Datasets}
	\label{tab7}
	\small
	\renewcommand\arraystretch{1.2}
	\begin{tabular}{m{0.15\linewidth}<{\centering}|m{0.14\linewidth}<{\centering}|m{0.14\linewidth}<{\centering}|m{0.14\linewidth}<{\centering} m{0.14\linewidth}<{\centering}}
		\hline \textbf{Data}  & \textbf{FB15K} & \textbf{WN11} & \textbf{FB13} \\
		\hline
		\hline \#Rel & 1,345 & 11 & 13 \\
		\hline \#Ent & 14,951 & 38,696 & 75,043 \\
		\hline \#Train & 483,142 & 112,581 & 316,232\\
		\hline \#Valid & 50,000 & 2,609 & 5,908\\
		\hline \#Test & 59,071 & 10,544 & 23,733\\
		\hline
	\end{tabular}
\end{table}

\textbf{Implementation.} We implemented TransE, TransH, TransR, TransG and ManifoldE for comparison, we directly reproduce the claimed results with the reported optimal parameters. The optimal settings of KSR are the learning factor $\alpha=0.01$, margin $\gamma=2.5$ and Laplace hyper-parameter $\sigma=0.04$. For a fair comparison within the same parameter quantity, we adopt three settings for dimensions: \textit{S1}($n=10, d=10$), \textit{S2}($n=20, d=10$) and \textit{S3}($n=90, d=10$), where $n$ denotes the number of knowledge views and $d$ indicates the number of semantic clusters for each view. We train the model until convergence but stop at most 2,000 rounds.

\subsection{Entity Classification}
\textbf{Motivation.} To test our semantics-specific performance, we conduct the entity classification prediction. Since the entity type such as \textit{Human Language}, \textit{Artist} and \textit{Book Author} represents some semantics-relevant sense, thus this task could justify KSR indeed addresses the semantic representation.

\textbf{Evaluation Protocol.} Overall, this is a multi-label classification task with 25/50/75 classes, which means for each entity, the method should provide a set of types rather one specific type. In the training process of the classifier, we adopt the concatenation of cluster distribution $([z_1|e], [z_1|e], ..., [z_n|e])$ as entity representation, where $[z_i|e]$ is a distribution implemented as a vector. For a fair comparison, our front-end classifier is identically the Logistic Regression in a one-versus-rest setting for multi-label classification. The evaluation is following \cite{neelakantan2015inferring}, which applies the mean average precision (MAP) that is commonly used in multi-label classification. Type@N means the task is involved with N types to be predicted.

\begin{table}
	\centering
	\caption{Evaluation results of Entity Classification.}
	\label{tab5}
	\begin{tabular}[t]{c|c|c|c}
		\hline \textbf{Metrics} & T@25 & T@50 & T@75 \\
		\hline 
		\hline Random & 39.5 & 30.5 & 26.0 \\
		\hline TransE & 82.7 & 77.3 & 74.2 \\
		\hline TransH & 82.2 & 71.5 & 71.4 \\
		\hline TransR & 82.4 & 76.8 & 73.6 \\
		\hline ManifoldE & 86.4 & 82.2 & 79.6 \\
		\hline
		\hline \textbf{KSR(\textit{S1})} & 90.7 & 85.6 & 83.3 \\
		\hline \textbf{KSR(\textit{S2})} & \textbf{91.4} & \textbf{87.6} & \textbf{85.1} \\
		\hline \textbf{KSR(\textit{S3})} & 90.2 & 86.1 & 83.1 \\
		\hline
	\end{tabular} 
\end{table}

\textbf{Results.} Evaluation results are reported in Tab.\ref{tab5}, noting that $S1,S2$ and $S3$ mean different settings for knowledge views and semantic clusters. We could observe that: 
\begin{enumerate}
	\item KSR outperforms all the baselines in a large margin, demonstrating the effectiveness of our model.
	\item Entity types represent some level of semantics, thus the better results illustrate that our method is indeed more semantics-specific.
\end{enumerate} 

\begin{table*}
	\centering
	\caption{Views with Significant Semantics in Semantic Analysis. Notably, \textit{No} corresponds to other meaningless or uninterpreted words, such as \textit{Is, The, Of, Lot, Good, Well, ...}. For each row, it is the view from first-level generation process and in each view (i.e. row), there list the three clusters with the significant words in this corresponding view.}
	\label{tab4}
	\begin{tabular}{c|c|c}
		\hline
		\textbf{No.} & \textbf{Views} & \textbf{Clusters for Different Views (Significant Words)} \\
		\hline
		\hline 1 & \textbf{\em Film} & \textbf{Yes} (Film, Director, Season, Writer), \textbf{Yes} (Awarded, Producer, Actor), \textbf{No}\\
		\hline 2 & \textbf{\em American} & \textbf{No, No, Yes} (United, States, Country, Population, Area) \\
		\hline 3 & \textbf{\em Sports} & \textbf{No, No, Yes} (Football, Club, League, Basketball, World Cup) \\
		\hline 4 & \textbf{\em Art} & \textbf{Yes}(Drama, Music, Voice, Acting), \textbf{Yes} (Film, Story, Screen Play), \textbf{No} \\
		\hline 5 & \textbf{\em Persons} & \textbf{No}, \textbf{Multiple} (Team, League, Roles), \textbf{Single}(She, Actress, Director, Singer) \\
		\hline 6 & \textbf{\em Location} & \textbf{Yes} (British, London, Canada, Europe, England), \textbf{No}, \textbf{No} \\
		\hline
	\end{tabular}
\end{table*}

\subsection{Knowledge Graph Completion}
\textbf{Motivation.} This task is a benchmark task, a.k.a ``Link Prediction'', which concerns the identification ability for triples. Many NLP tasks could benefit from this task, such as relation extraction \cite{hoffmann2011knowledge}.

\textbf{Evaluation Protocol.} We adopt the same protocol used in previous studies. For each testing triple $(h,r,t)$, we corrupt it by replacing the tail $t$ (or the head $h$) with every entity $e$ in the knowledge graph and calculate a probabilistic score of this corrupted triple $(h,r,e)$ (or $(e,r,t)$) with the score function $f_r(h,e)$ (or $f_r(e, t)$). After ranking these scores in descending order, we obtain the rank of the original triple. There are two metrics for evaluation: the mean reciprocal rank (MRR) and the proportion of testing triples whose ranks are not larger than 10 (HITS@10). This is called ``Raw'' setting. When we filter out the corrupted triples that exist in the training, validation, or test datasets, this is the ``Filter'' setting. If a corrupted triple exists in the knowledge graph, ranking it ahead the original triple is also acceptable. To eliminate this case, the ``Filter'' setting is preferred. In both settings, a higher MRR and HITS@10 mean better performance.

\textbf{Results.} Evaluation results are reported in Tab.\ref{tab2}. The parameter scale of the first segment is the least and that of the last segment is the most. We could observe that:
\begin{enumerate}
	\item  KSR outperforms all the baselines substantially, justifying the effectiveness of our model. Theoretically, the effectiveness originates from the semantics-specific modeling of KSR.
	\item Within the same parameter scale (i.e., \textit{the number of total parameters in these models are comparable}), compared to TransE, KSR improves 6.5\% relatively, while compared to TransR, KSR improves 19.5\%. The comparison illustrates KSR benefits from high-dimensional settings on knowledge views and clusters.
\end{enumerate} 

\begin{table}
	\centering
	\caption{Evaluation results of Knowledge Graph Completion. We order and segment the methods by parameter scalability. The parameter scale of the first segment is the least and that of the last segment is the most.}
	\label{tab2}
	\begin{tabular}[t]{c|c|c|c|c}
		\hline \textbf{FB15K} & \multicolumn{2}{c|}{\textbf{MRR (Filter)}} & \multicolumn{2}{c}{\textbf{HITS@10}(\%)} \\ 
		\hline Methods & Head & Tail & Raw & Filter \\
		\hline 
		\hline TransE & 35.6 & 40.1 & 48.5 & 66.1 \\
		\hline TransH  & 33.9 & 39.1 & 45.7 & 64.4 \\
		\hline \textbf{KSR(\textit{S1})} & \textbf{36.5} & \textbf{42.6} & \textbf{51.2} & \textbf{70.4} \\
		\hline
		\hline HOLE & - & - & - & 73.9 \\
		\hline \textbf{KSR(\textit{S2})} & \textbf{37.8} & \textbf{44.2} & \textbf{52.6} & \textbf{75.8} \\
		\hline
		\hline TransR & 25.1 & 30.0 & 48.2 & 68.7 \\
		\hline KG2E & - & - & 47.5 & 71.5 \\
		\hline TransG & - & - & 53.1 & 79.7 \\
		\hline ManifoldE & 34.2 & 40.0 & 52.1 & 79.8 \\
		\hline \textbf{KSR(\textit{S3})} & \textbf{40.0} & \textbf{45.5} & \textbf{52.9} & \textbf{82.1} \\
		\hline 
	\end{tabular}
\end{table}

\subsection{Triple Classification}
\textbf{Motivation.} In order to test the discriminative capability between true and false facts, triple classification is conducted. This is a classical task in knowledge graph embedding, which aims at predicting whether a given triple $(h,r, t)$ is correct or not. WN11 and FB13 are the benchmark datasets for this task. Note that the evaluation of classification needs negative samples, and the datasets have already provided negative triples.

\textbf{Evaluation Protocol.} The decision process is very simple as follows: for a triple $(h,r,t)$, if $f_r(h,t)$ is below a threshold $\sigma_r$, then positive; otherwise negative. The thresholds $\{\sigma_r\}$ are determined on the validation dataset.

\textbf{Results.} Accuracies are reported in Tab.\ref{tab9}. The following are our observations:
\begin{enumerate}
	\item KSR outperforms all the baselines remarkably. Compared to TransR, KSR (S3) improves by 1.8\% on WN11 and 5.9\% on FB13, which illustrates the effectiveness of our model.
	\item KSR (S3) outperforms other settings, which means our model benefits from high-dimensional settings.
\end{enumerate}

\subsection{Semantic Analysis: Case Study}
We conduct a case study to analyze the semantics of our model. For brevity, we explore the FB15K datasets with KSR ($n=10, d=3$), which employs 10 knowledge views and for each view assigns 3 clusters. In fact, FB15K is more complex to approach than this setting, thus many minor views and clusters have to be suppressed. The consideration of this setting is to facilitate visualization presentation. 

First, we analyze the specific semantics of each view. We leverage the entity descriptions to calculate the joint probability by the corresponding occurrence number of word $w$ in the textual descriptions of an entity $e$ and the inferred view-cluster $S_{e,i}$ of that entity. Therefore, we have:
\begin{eqnarray}
& [w^{=j}, z_i^{=c}] & \propto \#\left\{\exists e \in E, w_j \in D_e \wedge S_{e,i}=c\right\} \nonumber \\
& & = \sum_{e \in E} \delta_{w_j \in D_e~and~S_{e,i}=c}
\end{eqnarray}
where $D_e$ is the set of words in the description of entity $e$, $\delta$ is the indicator symbol and regarding $S_{e,i}$ the reader could refer to Section 3.1.

Then, we list the significant words in each cluster for each view. In this way, the semantics of views and clusters could be explicitly interpreted. We directly list the results in Tab.\ref{tab4}. There are six significant views, which are presented with clusters and significant words as evidence. This result strongly justifies our motivation of KSR. \textit{\textbf{Thus, with the significant words of the views and clusters, we can manually define the semantics of these views and clusters.}} Notably, the other four views are too vague to be recognized, because KSR is a latent space method similar to LDA \cite{blei2012probabilistic}.

Last, we present the semantic representations for three entities of different types: Film, Sport and Person. In the representations, the views are listed as \textit{Film, American, Sports, Person, Location, Drama}, while the clusters for the corresponding view are followed after the colon. For example, \textit{``Related''} is one cluster under the view of Film. \textit{\textbf{We list the views and the corresponding clusters in Tab.\ref{tab4}.}} For the view of person, the cluster could be \textit{``Unrelated''} such as for the entity \textit{China}, while the cluster could be single (e.g. for \textit{Johnathan Glickman}) or multiple (e.g. for \textit{Football Club Illichivets Mariupol}). We achieve the views and clusters by KSR, and then label the word meanings for the views and clusters by the entity descriptions, in above procedure. \textit{\textbf{Thus, with the word meaning of the views and clusters, we can manually define the semantics of these views and clusters.}}

\begin{enumerate}
	\item \textit{(Star Trek) = (Film: Related, American: Related, Sports: Unrelated, Person: Unrelated, Location: Unrelated, Drama: Related)}.\\ \textit{Star Trek} is the television series produced in America. Thus our semantic representations are quite coherent to the semantics of the entity.
	\item  \textit{(Football Club Illichivets Mariupol) = (Film: Unrelated, American: Unrelated, Sports: Related, Art: Unrelated, Persons: Multiple, Location: Related)}.\\ Its textual description is ``Football Club Illichivets Mariupol is a Ukrainian professional football club based in Mariupol'', which is accordant with the semantic representation. Note that, football club as a team is composed by multiple persons, which is the reason for \textit{Person: Multiple}.
	\item  \textit{(Johnathan Glickman)=(Film: Related, American: Unrelated, Sports: Unrelated, Art: Unrelated, Person: Single, Location: Unrelated)}. \\This person is a film producer, while we could not search out any nationality information about this person, but our semantic representation could still be interpretable.
\end{enumerate}

Finally, we also present the semantic representations for relation. 
\begin{enumerate}
	\item \textit{(Country Capital) = (Film: Unrelated, American: Unrelated, Sports: Unrelated, Art: Unrelated, Person: Unrelated, Location: Related)}. \\ As a common sense, a capital is a location, not sports or art, thus our semantic representations are reasonable.
\end{enumerate}

\begin{table}
	\centering
	\caption{Triple classification: accuracy(\%) for different embedding methods.}
	\label{tab9}
	\begin{tabular}{c|c|c|c}
		\hline \textbf{Methods} & \textbf{WN11} & \textbf{FB13} & \textbf{AVG.} \\
		\hline
		\hline SE & 53.0 & 75.2 & 64.1 \\
		\hline SME(bilinear) & 70.0 & 63.7 & 66.9\\
		\hline LFM  & 73.8 & 84.3 & 79.0\\
		\hline NTN  & 70.4 & 87.1 & 78.8 \\
		\hline TransE & 75.9 & 81.5 & 78.7 \\
		\hline TransH & 78.8 & 83.3 & 81.1 \\
		\hline TransR & 85.9 & 82.5 & 84.2\\
		\hline CTransR & 85.7 & N/A & N/A \\
		\hline KG2E & 85.4 & 85.3 & 85.4 \\
		\hline TransA & 83.2 & 85.4 & 84.3 \\
		\hline TransG & 87.4 & 87.3 & 87.4 \\
		\hline
		\hline KSR (S1) & 87.3 & 87.2 & 87.3 \\
		\hline KSR (S2) & 87.1 & 87.3 & 87.2 \\
		\hline KSR (S3) & \textbf{87.5} & \textbf{87.4} & \textbf{87.5}\\
		\hline
	\end{tabular}
\end{table}

\subsection{Semantic Analysis: Statistic Justification}
We conduct statistical analysis in the same setting as the previous subsection.

Firstly, we randomly select 100 entities and manually check out the correctness of semantic representations by common knowledge. There are 68 entities, the semantic representations for which are totally correct and also 19 entities, the representations for which are incorrect at only one view. There are just 13 entities in which the corresponding representations are incorrect at more than one view. Thus, the result proves the strong semantic expressive ability of KSR.

\begin{figure}
	\centering
	\label{fig:fig4}
	\includegraphics[width=0.8\linewidth]{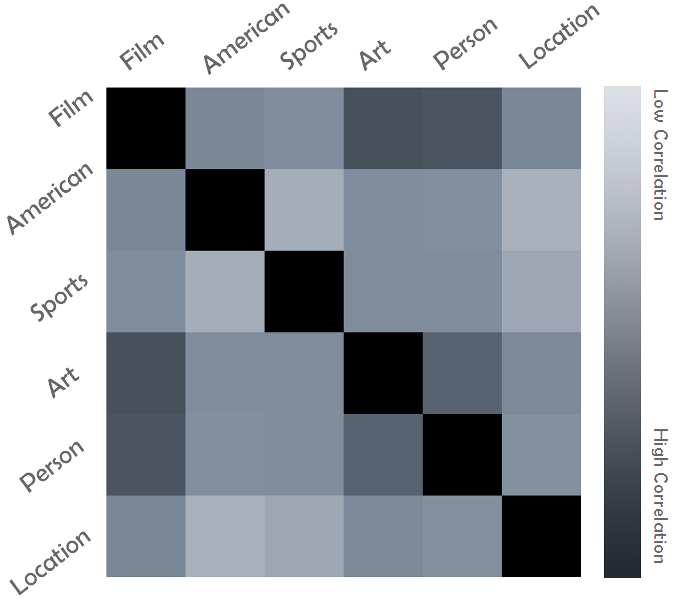}
	\caption{The heatmap of correlations between knowledge views in KSR. Darker color indicates higher correlation.}
\end{figure}

Secondly, if two views (both with cluster \textit{Yes}) co-occur in a semantic representation of an entity/relation, this knowledge element (entity/relation) contributes to the correlation between the two views. We make a statistics of the correlation and draw a heatmap in Fig.4, where the darker color corresponds to higher correlation. Looking into the details, those \textit{Sports:Related} entities would distribute all over the world, so they are almost \textit{American:Unrelated}. The result shows that the correlation between the two views is loose. \textit{Film} is highly correlated with \textit{Art} and \textit{Person}, which is accordant with our common knowledge. \textit{Location} indicates the geographical position outside U.S., thus it is loosely related to \textit{America}. 

\subsection{Entity Retrieval: Statistic Justification}
We conduct experiments for testing baselines and three composition methods in Section 4. We leverage KSR (S3) as our standard setting for this subsection. Firstly, we construct a dataset of 217 question-entity pairs. For an exhaustive test, we choose the sentences with different lengths in different aspects. There list some exemplified data items:
\begin{itemize}
	\item \textit{Moive-Related Long Query:} \\ An American 2011 biographical sports drama lm directed by Bennett Miller from a screenplay by Steven Zaillian and Aaron Sorkin.
	\item \textit{Figure-Related Long Query:} \\ A man who is a contemporary writer, playwright, screenwriter, actor and movie director in Kannada language. His rise as a playwright in 1960s, marked the coming of age of Modern Indian playwriting in Kannada, just as Badal Sarkar did in Bengali, Vijay Tendulkar in Marathi, and Mohan Rakesh in Hindi.
	\item \textit{Science-Related Short Query:} \\ Who proposes the Relativity Theory?
	\item \textit{Location-Related Short Query:} \\ Which provinces are the neighbor of Beijing?
\end{itemize}  

We choose two baselines: Text Classification and Information Retrieval. Text Classification treats each entity description as a category and classifies the query into one of the categories to retrieve the entity with the method of Support Vector Machine (SVM). Information Retrieval treats the entity descriptions as documents and takes advantage of traditional information retrieval method to search the answer document/entity for the query. \textit{Notably, conducting entity retrieval based on entity description is novel, thus there is no other suitable baselines.}

We leverage the accuracy to metric the performance. However, regarding the queries with more than one entity as answers, if one of the answer entities is retrieved, the question is tackled. In detail, entity corresponds to entity description as document in IR or category in SVM. Thus, if IR achieves the document according to the query, we can obtain the entity corresponded to the document (i.e. entity description). Or if SVM classifies the query to the category we can obtain the entity corresponded to the category (i.e. entity description). For our methods, please refer to Section 4. \textbf{\textit{By the metric of accuracy, we mean HITS@1.}}

\begin{table}
	\centering
	\caption{Accuracy (\%) of Entity Retrieval}
	\label{tab11}

	\begin{tabular}{c|c}
		\hline \textbf{Methods} & \textbf{Accuracy} \\ 
		\hline
		\hline Text Classification & 39.1 \\
		\hline Information Retrieval & 78.5 \\
		\hline
		\hline KSR with Average Pooling & 86.6 \\
		\hline KSR with Naive Bayes Composition & \textbf{91.7} \\
		\hline KSR with LSTM & 85.7 \\
		\hline 
	\end{tabular} 
\end{table}

Accuracies are reported in Tab.\ref{tab11}. The following are our observations:
\begin{enumerate}
	\item KSR beats all the baselines, verifying the effectiveness and semantic modeling ability of our model. 
	\item Naive Bayes Composition takes advantage of Bayesian inference to outperform Average Pooling.
	\item Since there are only at most 15,000 entities with textual descriptions, LSTM needs more data to boost the performance.
\end{enumerate}

\subsection{Entity Retrieval:  Case Study}
There are three sub-tasks: single factoid, multiple factoid and inferential query. Single factoid query means there is only one entity for the factoid question, while multiple factoid query means there are multiple entities for the factoid question. Inferential query needs logic inference or language comprehension ability to answer the question. Notably, all the results of this subsection are obtained by KSR with Naive Bayes Composition and we only list the top answer entities.

\textit{\textbf{Notably, all the results presented in this subsection are the top 1 results of our proposed model.}}

\subsubsection{Single Factoid Query}
There list some single factoid cases:
\begin{itemize}
	\item An American 2011 biographical sports drama lm directed by Bennett Miller from a screenplay by Steven Zaillian and Aaron Sorkin. \\ \textbf{Money Ball}
	\item The social science of human social behavior and its origins, development, organizations, and institutions. \\ \textbf{Sociology}
	\item A man who is a contemporary writer, playwright, screenwriter, actor and movie director in Kannada language. His rise as a playwright in 1960s, marked the coming of age of Modern Indian playwriting in Kannada, just as Badal Sarkar did in Bengali, Vijay Tendulkar in Marathi, and Mohan Rakesh in Hindi. \\ \textbf{Girish Raghunath Karnad}
	\item A professional baseball team located in Chicago, Illinois, USA. \\ \textbf{Chicago Club}
	\item Who proposes the Relativity Theory? \\ \textbf{Albert Einstein}
	\item Who proposes the Hawking Radio? \\ \textbf{Stephen William Hawking}
	\item Which company is best at chips? \\ \textbf{Intel Corporation}
	\item Which company is best at Multi-Media? \\ \textbf{Adobe Systems Incorporated}
	\item Which company is best at selling operating system? \\ \textbf{Microsoft Corporation}
	\item Which company is best at selling sport shoes? \\ \textbf{Nike, Inc.}
	\item What is the Chinese province with capital Taiyuan? \\ \textbf{Shan-Xi}
	\item What is the Chinese province with capital Changsha? \\ \textbf{Hu-Nan}
\end{itemize}
From the cases, we could conclude that our model is very effective for answering single factoid query. The results of entity retrieval demonstrate the potentials of semantic representations, which are founded on the effectiveness of KSR.

\subsubsection{Multiple Factoid Query}
There list some multiple factoid cases:
\begin{itemize}
	\item Which provinces are the neighbor of Beijing? \\ \textbf{Tianjin} \\ \textbf{He-Bei}
	\item Which countries are the neighbor of China? \\ \textbf{Burma} \\ \textbf{Vietnam} \\ \textbf{India}
\end{itemize}
The results of this subsubsection illustrate the possibilities of semantic representations for jointing knowledge and language, which are founded on the effectiveness of KSR.  

\subsubsection{Inferential Query}
There list two inferential queries:
\begin{itemize}
	\item The people in which Chinese province is richest? \\ \textbf{Macao} \\ \textbf{Su-Zhou} (Wrong Answer, Because Su-Zhou is a city rather than a province) \\ \textbf{Hong Kong} \\ \textbf{Jiang-Su}
	\item Which universities are famous in Kyoto? \\ \textbf{Kyoto University}
\end{itemize}
The results demonstrate KSR could infer complex language problems, not limited to simple factoid questions. 


\subsection{Discussion \& Error Analysis}
We also test our model by discarding the first-level, making our model single-view clustering. In this way, the settings should be $(n = 1, d = 10)$. We process the clusters achieved in the same procedure as Section 5.5. Then, we analyze the significant words in each cluster. Only one cluster may be related to \textit{Film}, because the top words for this one are listed as \textit{Good, Director, Film, States, Writer, Of, etc}. All the other nine clusters are mixed with multiple semantics which are hard to distinguish.

We also note some negative samples for entity retrieval. 
\begin{itemize}
	\item Which provinces are the neighbors of Tianjin? \\
	\textbf{Tianjin} (Wrong Answer) \\
	\textbf{Beijing} \\
	\textbf{He-Bei}
	\item Which cities are the neighbors of Kyoto? \\
	\textbf{Kyoto} (Wrong Answer)
\end{itemize}  

The first negative sample shows the semantic composition by KSR with Naive Bayes will be affected by the keyword. If the keyword is an entity in the knowledge graph, the first achieved answer may be the keyword rather than the true answer. For this flaw, we may design logic rules to enhance our model. The second negative sample shows the limit of our knowledge graph, which means if the answer entity is out of our knowledge graph, we can not obtain the corresponding answer. Thus, to employ stronger and larger dataset is necessary, which proposes the issue of algorithm efficiency.

\section{Semantic Principle}
In this section, we propose \textbf{\textit{Semantic Principle}} that cluster equals semantics by definition, mathematically:
\begin{equation}
	\boxed{Cluster \doteq Semantics}
\end{equation}
In other words,we call the semantics equaling to cluster as \textbf{\textit{clustering semantics}}. For the example of entities in knowledge graph, all the Beijing-related entities group into the Beijing-related cluster, thus all the entities in this cluster are semantically related to Beijing. For another example of classification in machine learning, the categories all lay as clusters with boundaries between each other. Intuitively, we prove semantic principle in conclusion.

\section{Conclusion}
Based on the multi-view clustering framework, we provide a novel model that Knowledge Semantic Representation (KSR), which is a two-level hierarchical generative process to semantically represent knowledge. This model is able to produce interpretable representations. We also evaluate our method with extensive studies. Experimental results justify the effectiveness and the capability of semantic expressiveness in our model.

We have already released our datasets and codes in Github \url{https://github.com/bookmanhan/Embedding}, which is a framework of knowledge graph embedding. This work is supported by Fujian Provincial Key Laboratory of Information Processing and Intelligent Control (Minjiang University) No.MJUKF-IPIC201804.



\bibliographystyle{IEEEtran}
\bibliography{KSR}
\vspace{-40pt}
\begin{IEEEbiography}
[{\includegraphics[width=1in,height=1.2in,clip,keepaspectratio]{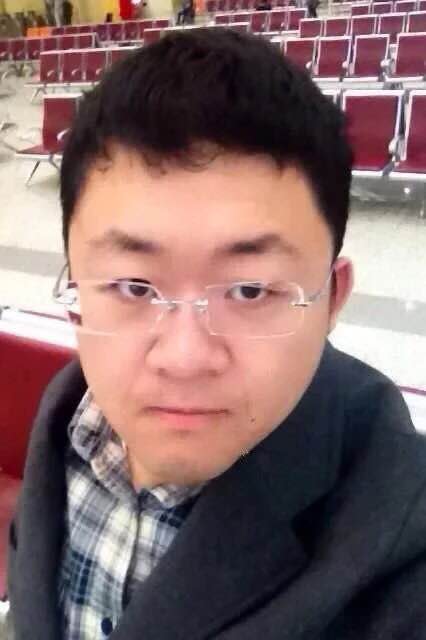}}]{Han Xiao} received his PH.D from Tsinghua University, Beijing, China as an outstanding graduate in 2017. He works at Xiamen University as an assistant professor. Currently, he is responsible for the course of artificial intelligence principles. His research interests focus on knowledge graph and recommendation systems, in which domains, several top conference and journal papers are published.
\end{IEEEbiography}
\vspace{-40pt}
\begin{IEEEbiography}
[{\includegraphics[width=1in,height=1.2in,clip,keepaspectratio]{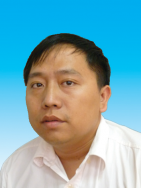}}]{Yidong Chen} received his PH.D from Xiamen University, Fujian, China in 2008. He works at Xiamen University as an associate professor. Currently, he is responsible for the course of machine translation. His research interests focus on machine translations.
\end{IEEEbiography}
\vspace{-40pt}
\begin{IEEEbiography}
[{\includegraphics[width=1in,height=1.2in,clip,keepaspectratio]{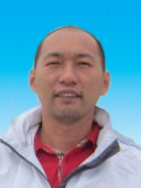}}]{Xiaodong Shi} received his PH.D from National Defense Technology University, Changsha, China in 1994. He works at Xiamen University as a professor. Currently, he is responsible for the course of advanced artificial intelligence. His research interests focus on natural language processing.
\end{IEEEbiography}

\end{document}